\documentclass[10pt, conference, compsocconf]{IEEEtran}

\usepackage[ruled,,vlined,lined,linesnumbered]{algorithm2e}
\usepackage{float}
\usepackage{subfigure}
\usepackage{amsmath}
\usepackage[pdftex]{graphicx}
\DeclareGraphicsExtensions{.png,.jpg}
\newtheorem{theorem}{Theorem}
\newtheorem{remark}{Remark}

\begin{document}

\title{A new stopping criterion for the mean shift iterative algorithm}

\author{\IEEEauthorblockN{Roberto Rodr\'iguez, Esley Torres, Yasel Garc\'es, Osvaldo Pereira}
\IEEEauthorblockA{Institute of Cybernetics, Mathematics and Physics (ICIMAF)\\
Digital Signal Processing Group\\
Havana, Cuba\\
Email: $\{$rrm,esley,ygarces,opereira$\}$@icimaf.cu}
\and
\IEEEauthorblockN{Juan H. Sossa}
\IEEEauthorblockA{National Polytechnic Institute (IPN)\\
Computing Research Center (CIC)\\
D.F., Mexico\\
Email: hsossa@cic.ipn.mx}
}

\maketitle

\begin{abstract}
The mean shift iterative algorithm was proposed in 2006, for using the entropy as a stopping criterion. From then on, a theoretical base has been developed and a group of applications has been carried out using this algorithm. This paper proposes a new stopping criterion for the mean shift iterative algorithm, where stopping threshold via entropy is used now, but in another way. Many segmentation experiments were carried out by utilizing standard images and it was verified that a better segmentation was reached, and that the algorithm had better stability. An analysis on the convergence, through a theorem, with the new stopping criterion was carried out. The goal of this paper is to compare the new stopping criterion with the old criterion. For this reason, the obtained results were not compared with other segmentation approaches, since with the old stopping criterion were previously carried out \cite{Comaniciu02}.
\end{abstract}


\IEEEpeerreviewmaketitle

\section{Introduction}

Image analysis is a scientific discipline providing theoretical foundations and methods for solving problems appearing in a range of areas as diverse as chemistry, physics, biology, geography, medicine, astronomy, robotics and industrial manufacturing. Besides traditional approaches based on continuous models, which require numeric computation, and which always involve the problem of rounding and approximation, ``combinatorial'' approaches to image analysis (also named ``digital'' or ``discrete'' approaches) have been developed during the last 60 years. These latter approaches are based on studying combinatorial properties of the digital data sets under consideration and generally providing useful algorithms for image analysis which are more efficient and accurate than those based on continuous models.

In order to cope with the variety of image processing and computer vision challenges, several techniques have been introduced and developed, quite often with great success. Among the different techniques that are currently in use, there are, for example, soft computing techniques. Soft computing is an emerging field that consists of complementary elements of fuzzy logic, neural computing, evolutionary computation, machine learning and statistical reasoning, and often offers solutions where conventional approaches fail.

Segmentation is the fundamental process which partitions a data space into meaningful salient regions. Image segmentation essentially affects the overall performance of any automated image analysis system. Thus, its quality is of the utmost importance. Image regions, homogeneous with regard to some usually statistical criterion or color measure, which result from a segmentation algorithm are analyzed in subsequent interpretation steps. Statistical criterion based image segmentation has been an area of intense research activity during the past forty years and many algorithms were published in consequence of all this effort, starting from simple thresholding methods up to the most sophisticated random field type methods. Unsupervised methods which do not assume any prior scene knowledge which can be learned to help segmentation processes are obviously more challenging than the supervised ones.

The mean shift (MSH) is a non-parametric procedure that has demonstrated to be an extremely versatile tool for feature analysis. It can provide reliable solutions for many computer vision tasks \cite{Comaniciu02}. The mean shift method was proposed in 1975 by Fukunaga and Hostetler \cite{Fukunaga75}. It was largely forgotten until Cheng's paper rekindled interest in it \cite{Cheng95}. Unsupervised segmentation by means of the mean shift method carries out as a first step a smoothing filter before segmentation is performed \cite{Comaniciu02}. Mean shift iterative algorithm was proposed in 2006 and this has been performed in many works by using the entropy as a stopping criterion \cite{Rodriguez11,Rodriguez11a,Rodriguez12,Rodriguez08,Dominguez11}.

The term of entropy is not a new concept in the field of information theory. Entropy has been used in image restoration, edge detection and recently as an objective evaluation method for image segmentation \cite{Zhang03}. The novelty of the proposed algorithm is the use of the entropy as a stopping criterion. The choice of entropy as a measure of goodness deserves several observations, which will be detailed in next section.

This paper proposes a new stopping criterion for the MSH iterative algorithm, where stopping threshold via entropy is used, now, in another way. Many segmentation experiments, by utilizing standard images, were carried out using this new stopping criterion. This paper compares the stability of MSH iterative algorithm using the new stopping criterion with regard to the old stopping criterion used in \cite{Rodriguez11,Rodriguez11a,Rodriguez12,Rodriguez08,Dominguez11}. Good segmentation was reached and the algorithm had better stability. An analysis on the convergence, through a theorem, with the new stopping criterion was carried out.

The remainder of the paper is as follows. In Section \ref{THEORICAL ASPECTS}, the more significant theoretical aspects of the mean shift and entropy are given. Section \ref{THE MEAN SHIFT ITERATIVE ALGORITHM} describes our MSH iterative algorithm with the old and new stopping criterion. In this section the theorem that ensures the convergence is proposed. The experimental results, comparisons and discussion are presented in Section \ref{EXPERIMENTAL RESULTS. ANALYSIS AND DISCUSSION}. Finally, in Section \ref{CONCLUSIONS} the conclusions are given.

\section{THEORICAL ASPECTS}
\label{THEORICAL ASPECTS}

\subsection{Mean Shift}
The basic concept of the mean shift algorithm is as follows: Let $x_{i}$ be an arbitrary set of $n$ points in the $d$ dimensional space. The kernel density estimation $f(x)$ is obtained by means of the kernel function $K(x)$ and window radius $h$. Function $f(x)$ is defined as
\begin{eqnarray}
f(x) =\frac{1}{nh^d}\sum\limits_{i=1}^{n}K\left( \frac{x-x_{i}}{h}\right) .
\end{eqnarray}

Here, the {\itshape Epanechnikov} function is chosen as the kernel function. The {\itshape Epanechnikov} function is defined as, 

\begin{align}
K_E(x)=\left\{
\begin{array}{ll}
\frac{1}{2} c^{-1}_{d} (d+2)\left( 1-\left\|x\right\|^2\right),  & \mbox{if}\ \left\|x\right\|<1\\
& \\
0, & \mbox{otherwise}.\\
\end{array}
\right.
\end{align}

The differential function $f(x)$ is formulated as

\begin{eqnarray}
\widehat{\nabla} f(x) = \nabla \widehat{f(x)} =\frac{1}{nh^d}\sum\limits_{i=1}^{n}\widehat K\left( \frac{x-x_{i}}{h}\right) ,
\end{eqnarray}

\begin{align}
\label{label4}
\widehat{\nabla}f_E(x)&=
\frac{1}{n(h^d c_d)}\frac{d+2}{h^2}\sum\limits_{x_i \in S_h(x)} (x_i-x)\nonumber\\
&=\frac{n_x}{n(h^d c_d)}\frac{d+2}{h^2}\frac{1}{n_x}\sum\limits_{x_i \in S_h(x)} (x_i-x),
\end{align}

where region  $S_{h}(x)$ is a hyper sphere of radius $h$ having volume $h^{d} c_{d}$, centred at $x$, and containing $n$ data points; that is, the uniform kernel. In addition, in this case $d=3$, for the $x$ vector of three dimensions, two for the spatial domain and one for the range domain (gray levels). The last factor in expression (\ref{label4}) is called the sample mean shift,

\begin{align}
\label{label5}
M_{h,U}(x)&=\frac{1}{n_x}\sum\limits_{x_i \in S_h(x)}(x_i-x)\nonumber\ \\ &=\left(\frac{1}{n_x}\sum\limits_{x_i \in S_h(x)}x_i\right)-x.
\end{align}

The quantity $\frac{n_x}{n(h^{d} c_{d})}$ is the kernel density estimate $\widehat{f_U}(x)$ (where $U$ means the uniform kernel) computed with the hyper sphere $S_{h}(x)$, and thus we can write the expression (\ref{label4}) as:

\begin{eqnarray}\label{label6}
\widehat{\nabla}f_{E}(x)=\widehat{f_{U}}(x)\frac{d+2}{h^2} M_{h,U}(x)
\end{eqnarray}
which yields,
\begin{eqnarray}\label{label7}
M_{h,U}(x)=\frac{h^2}{d+2}\frac{\widehat{\nabla}f_E(x)}{\widehat{f_{U}}(x)}.
\end{eqnarray}

Equation (\ref{label7}) shows that an estimate of the normalized gradient can be obtained by computing the sample mean shift in a uniform kernel centered on $x$. In addition, the mean shift has the gradient direction of the density estimate at $x$ when this estimate is obtained with the {\itshape Epanechnikov kernel}. Since the mean shift vector always points towards the direction of the maximum density increase, it can define a path leading to a local density maximum; that is, to the density mode.

In \cite{Comaniciu00}, it was proved that the obtained {\itshape mean shift procedure} by the following steps, guarantees the convergence:
	
\begin{itemize}
\item computing the mean shift vector $M_h(x)$
\item	translating the window $S_h(x)$ by $M_h(x)$
\end{itemize}

Therefore, if the individual mean shift procedure is guaranteed to converge, a recursively procedure of the mean shift also converges. Other related works with this issue can be seen in \cite{Suyash06,Grenier06}. 

\subsection{Entropy}
From the point of view of digital image processing the entropy of an image is defined as:
\begin{eqnarray}
\label{entropy}
E=-\sum\limits_{x=0}^{2^B-1}p_{x} \log_2{p_{x}},
\end{eqnarray}
where $B$ is the total quantity of bits of the digitized image  and by  agreement $\log_2(0)=0$ ; $p(x)$ is the probability of occurrence of a gray-level value. Within a totally uniform region, entropy reaches the minimum value. Theoretically speaking, the probability of occurrence of the gray-level value, within a uniform region is always one. In practice, when one works with real images the entropy value does not reach, in general, the zero value. This is due to the existent noise in the image. Therefore, if we consider entropy as a measure of the disorder within a system, it could be used as a good stopping criterion for an iterative process, by using the mean shift iterative algorithm. More goodness on entropy can be seen in \cite{Rodriguez11a,Zhang03}.

\section{THE MEAN SHIFT ITERATIVE ALGORITHM}
\label{THE MEAN SHIFT ITERATIVE ALGORITHM}
The Mean Shift Iterative Algorithm (MSHIA) with the old stopping criterion is composed of the following steps. Let $ent1$ be the initial value of the entropy of the first iteration. Let $ent2$ be the second value of the entropy after the first iteration. Let $errabs$ be the absolute value of the difference of entropy between the first and the second iteration. Let $edsEnt$ be the threshold to stop the iterations; that is, to stop when the relative rate of change of the entropy from one iteration to the next, falls below this threshold. Then, the segmentation algorithm comprises the following steps:

\begin{algorithm}[h]
\caption{Algorithm with the Old Stopping Criterion}\label{AlgoritmoOld}
	Initialize $ent2$, $errabs$ and $edsEnt$\; 
	While $errabs > edsEnt$, then\;
	Filter the image according to the mean shift algorithm; store in $Z[k]$ the filtered image\;
	Calculate the entropy from the filtered image according to expression (\ref{entropy}); store in ent1\;
	Calculate the absolute difference with the entropy value obtained in the previous step; $errabs = \vert ent1 - ent2 \vert$\;
	Update the value of the parameters an image; $ent2 = ent1$; $Z[k +1] = Z[k]$\;
\end{algorithm}

\begin{algorithm}[h]
\caption{The Algorithm with the New Stopping Criterion}\label{AlgoritmoNew}
 Initialize $edsEnt$ and $B1$, let $B1$ be equal to original image\;	
 While $errabs > edsEnt$, then\;
 Filter the original image according to the mean shift algorithm; store in $B2$ the filtered image\;
 Calculate the absolute difference with the image obtained in the previous step; $C = \vert B2 - B1\vert$\;
 Calculate the entropy to the previous image; store in $errabs$\;
 Update the images: $B1 = B2$\;
\end{algorithm}

One can observe that with the new stopping criterion the entropy is used in another way. This paper will prove that with the new stopping criterion the MSHIA offers greater stability. Moreover, the following theorem resulted very interesting.

\begin{theorem} 
\label{theorem1}
When the entropy of the absolute difference between the iteration image and the following is taken as a stopping criterion in the mean shift iterative algorithm, whatever the chosen threshold this is enough condition to achieve the convergence. In addition, at the limit the entropy is zero.
\end{theorem}

This theorem ensures the convergence of the MSHIA with the new stopping criterion and determines what happens with entropy at the limit. The proof of this theorem can be found in the appendix.

\section{EXPERIMENTAL RESULTS. ANALYSIS AND DISCUSSION}
\label{EXPERIMENTAL RESULTS. ANALYSIS AND DISCUSSION}

All segmentation experiments were carried out by using a uniform kernel. In order to be effective the comparison between the old stopping criterion and the new stopping criterion, the same value of $h_r$ and $h_s$ in the MSHIA were used. The principal goal of this section is to evaluate the new stop criterion in the MSHIA and to prove the stability of the algorithm with regard to the old stopping criterion. For this reason, comparisons with other segmentation approaches will not be carried out. In \cite{Rodriguez11a} were compared the obtained results with the MSHIA through the old stopping criterion with other segmentation methods.

\begin{figure}[ht]
	\centering
		\subfigure[Astro]{\label{Figura1a}\includegraphics[width=2.733cm, height=2.75cm]{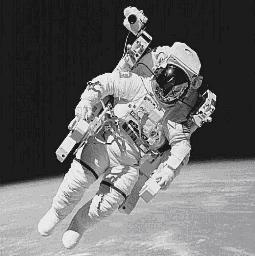} }\hspace{-0.1cm}
		\subfigure[Old criterion]{\label{Figura1b}\includegraphics[width=2.75cm, height=2.75cm]{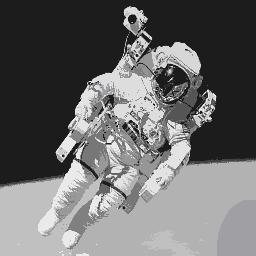}}
		\subfigure[New criterion]{\label{Figura1c}\includegraphics[width=2.75cm, height=2.75cm]{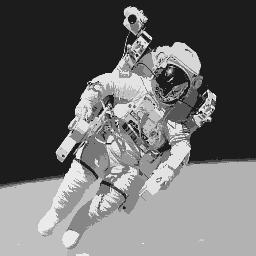}}
		\subfigure[Barbara]{\label{Figura1d}\includegraphics[width=2.75cm, height=2.75cm]{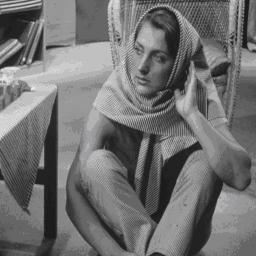}}
		\subfigure[Old criterion]{\label{Figura1e}\includegraphics[width=2.75cm, height=2.75cm]{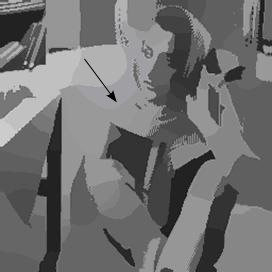}}
		\subfigure[New criterion]{\label{Figura1f}\includegraphics[width=2.75cm, height=2.75cm]{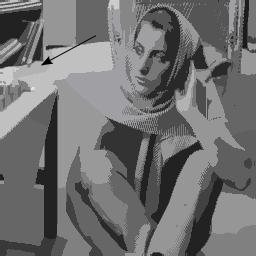}}
		\subfigure[Cameraman]{\label{Figura1g}\includegraphics[width=2.75cm, height=2.75cm]{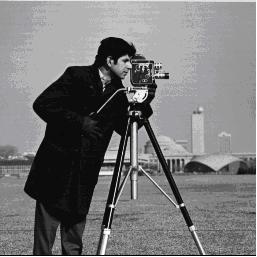}}
		\subfigure[Old criterion]{\label{Figura1h}\includegraphics[width=2.75cm, height=2.75cm]{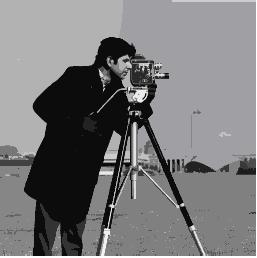}}
		\subfigure[New criterion]{\label{Figura1i}\includegraphics[width=2.75cm, height=2.75cm]{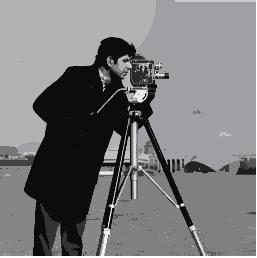}}
		\subfigure[Lena]{\label{Figura1j}\includegraphics[width=2.75cm, height=2.75cm]{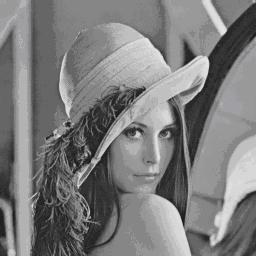}}
		\subfigure[Old criterion]{\label{Figura1k}\includegraphics[width=2.75cm, height=2.75cm]{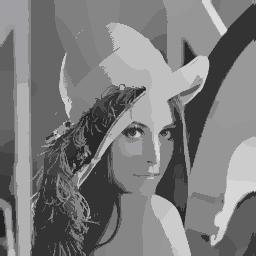}}
		\subfigure[New criterion]{\label{Figura1l}\includegraphics[width=2.75cm, height=2.75cm]{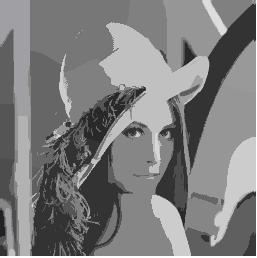}}

	\caption{Used images for segmentation. In the first column are shown the original images; in the second, the segmentation using the old stopping criterion and the third column shows the segmented images using the new stopping criterion.}
	\label{fig:segmentation}
\end{figure}

Figure \ref{fig:segmentation} shows the segmentation of the used images. Observe that, in all cases, the iterative mean shift algorithm had better results when using the new stopping criterion.

When one compares the segmented images with the old and new criterion it can be observed that with the new stopping criterion (see, for example, central images in Figure \ref{fig:segmentation}), more defined homogeneous zones were obtained. It can be seen that with the old stopping criterion the segmentation gave regions where different gray levels were originated. However, these regions really should have only one gray level. In other zones happened the opposite (see arrow in Figure \ref{Figura1f}). For example, Figure \ref{Figura1e} shows (visually) that the segmentation, with the old stopping criterion, is more diluted (see arrow). However, in Figure \ref{Figura1f} the segmentation with the new stopping criterion has the zones more delimited. One can find this same result when observing the other segmented images. This is one of the principal advantages of the new stopping criterion with regard to the old criterion. These good results are obtained because the defined new stopping criterion through the natural distance between images offers greater stability to the mean shift iterative algorithm.

\begin{figure}[ht]
	\centering
		\subfigure[Old criterion]{\includegraphics[width=4.2cm, height=4.4cm]{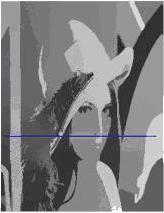}}
	  \subfigure[Profile]{\includegraphics[width=4.2cm, height=4.3cm]{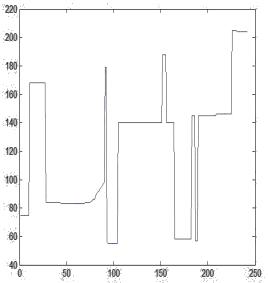}}
	  \subfigure[New criterion]{\includegraphics[width=4.2cm, height=4.4cm]{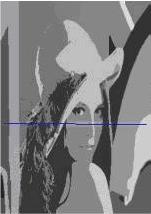}}
		\subfigure[Profile]{\includegraphics[width=4.2cm, height=4.3cm]{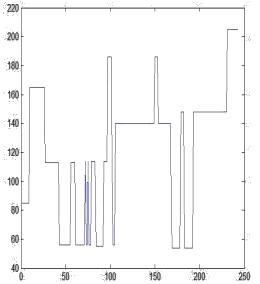}}
		
\caption{Intensity profiles through the segmented images. Profiles are indicated by lines. First two columns are the segmented images with the old stopping criterion and its profile. The last columns are the segmented images with the new stopping criterion and its profile.}
\label{fig:Profile}
\end{figure}

Figure \ref{fig:Profile} shows the profiles of the obtained segmented images by using the two stopping criteria.

\begin{figure}[ht]
	\centering
		\subfigure[Old criterion]{\includegraphics[width=4.2cm, height=4.3cm]{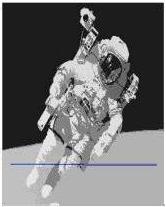}}
	  \subfigure[Profile]{\includegraphics[width=4.2cm, height=4.3cm]{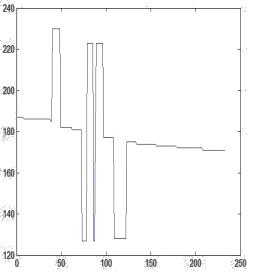}}
	 \subfigure[New criterion]{\includegraphics[width=4.2cm, height=4.3cm]{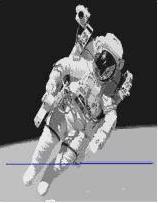}}
		\subfigure[Profile]{\includegraphics[width=4.2cm, height=4.3cm]{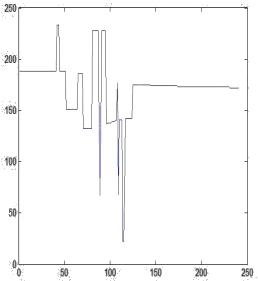}}
\end{figure}

\begin{figure}[ht]
	\centering
		\subfigure[Old criterion]{\includegraphics[width=4.2cm, height=4.2cm]{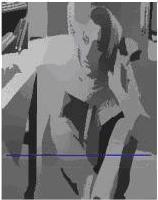}}
	  \subfigure[Profile]{\includegraphics[width=4.2cm, height=4.2cm]{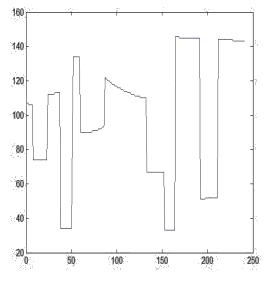}}
	 \subfigure[New criterion]{\includegraphics[width=4.2cm, height=4.2cm]{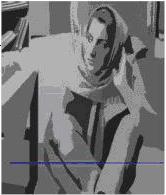}}
		\subfigure[Profile]{\includegraphics[width=4.2cm, height=4.2cm]{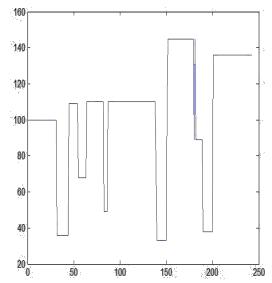}}
\end{figure}

\begin{figure}[ht]
	\centering
		\subfigure[Old criterion]{\includegraphics[width=4.1cm, height=4.1cm]{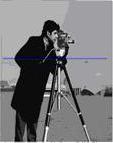}}
	  \subfigure[Profile]{\includegraphics[width=4.1cm, height=4.1cm]{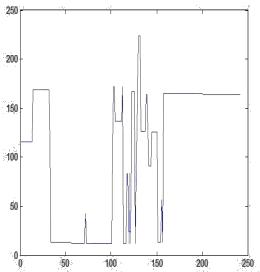}}
	 \subfigure[New criterion]{\includegraphics[width=4.1cm, height=4.1cm]{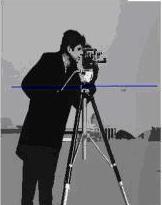}}
		\subfigure[Profile]{\includegraphics[width=4.1cm, height=4.1cm]{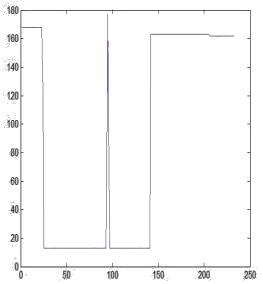}}
\end{figure}

The plates that appear in Figures \ref{fig:Profile} of the second and the last columns are indicative of equal intensity levels. In both graphics the abrupt falls of intensities to others represent different regions in the segmented images. Note, for example, that in Figures where the segmented image appears with the new stopping criterion there are, in the same region of the segmentation, least variation of the pixel intensities with regard to the segmented images with the old stopping criterion. This illustrates that, in these cases the segmentation was better when the new stopping criterion was used.

\begin{figure}[ht]
	\centering
		\subfigure[Barbara]{\includegraphics[width=2.72cm, height=2.72cm]{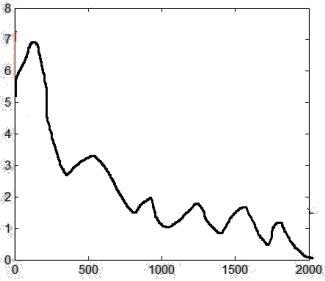}}
	  \subfigure[Cameraman]{\includegraphics[width=2.72cm, height=2.72cm]{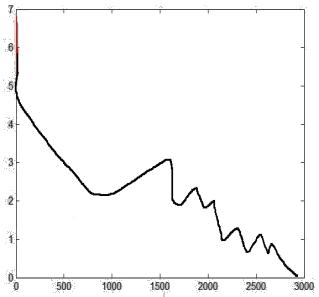}}
	  \subfigure[Lena]{\includegraphics[width=2.72cm, height=2.72cm]{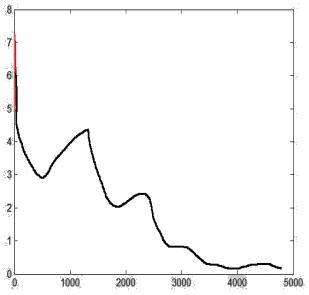}}
		\subfigure[Barbara]{\includegraphics[width=2.72cm, height=2.72cm]{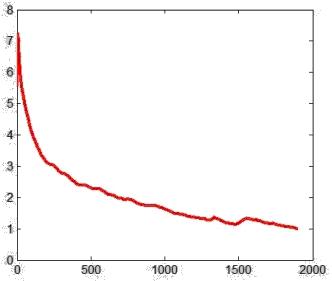}}
	  \subfigure[Cameraman]{\includegraphics[width=2.72cm, height=2.72cm]{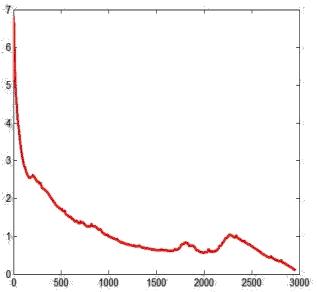}}
	  \subfigure[Lena]{\includegraphics[width=2.72cm, height=2.72cm]{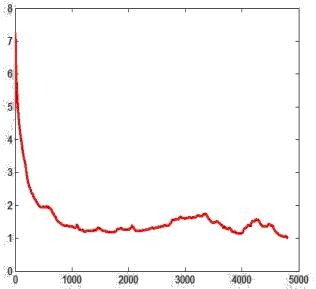}}
	  \caption{Stopping criterion. In the first row appears the performance of the old stopping criterion, while in the second row the performance of the new stopping criterion is shown.}
\label{Iteration graphs}
\end{figure}

Figure \ref{Iteration graphs} shows three examples of the performance of the two stopping criterion in the experimental results. In the ``x'' axis appears the iterations of the MSHIA and in the ``y'' axis the obtained values by the stopping criterion in each iteration of the algorithm are shown.

The graphics of iterations of the new stopping criterion (second row) show a smooth behavior; that is, the stopping criterion has a better stable performance through the MSHIA. This is the goal of the presentation of these graphics, to observe the oscillations in the ``y'' axis (stability). The new stopping criterion not only has good theoretical properties (see theorem), but also, in practice, has very good behavior. When one analyzes the performance of the old stopping criterion in the experimental results (first row), one can see that the performance in the MSHIA is unstable. In general, this can originate unsuitable segmented images.

\section{CONCLUSIONS}
\label{CONCLUSIONS}

In this work, a new stopping criterion for the MSHIA was proposed. The new stopping criterion is based on obtaining the entropy of the absolute difference between two images. In such sense, a theorem for the convergence was proposed. It was proven, through many experiments by using standard images, that the new stopping criterion had very good performance in the MSHIA. In addition, this was more stable that the old criterion.

\section*{Acknowledgment}
H. Sossa thanks SIP-IPN and CONACYT for the economical supports under grants 20131182 and 155014, respectively. We all thank the reviewers for their comments on the improvement of this paper.

\appendix[Proof of Theorem \ref{theorem1}]
Firstly, proof by absurd the last part of the theorem. It is known that;
\setcounter{equation}{0}
\begin{eqnarray}\label{entropy1}
E=-\sum\limits_{x=0}^{2^B-1}p_{x} \log_2{p_{x}},
\end{eqnarray}

where $B$ is the total quantity of bits of the digitized image and by agreement $\log_{2}(0)=0$; $p(x)$ is the probability of occurrence of a gray-level value.

When developing expression (\ref{entropy1}), 
\begin{align}
\label{entropy2}
E&=-\sum\limits_{x=0}^{2^B-1}p_{x} \log_2{p_{x}}\nonumber\\
&=-p_{0}\log_2{p_{0}}- \ldots -p_{2^B-1} \log_2{p_{2^B-1}}.
\end{align}

\vfill\break
Let $E\neq 0 \Rightarrow \forall i\ p_i\neq 1,\ p_i\neq 0$.

However, in expression (\ref{entropy2}) one only has interest in the gray level of set $G = \left\lbrace j | p_j \neq 0\right\rbrace $, since the gray level equal to zero does not contribute to entropy.
Let $I_k$ and $I_{k+1}$ be the resulting images of the iterations $k$ and $k+1$, and let $G_i = |I_{k+1} - I_k|$ be, $k = 1, 2, 3,\ldots$.

Let $E_i$ be the corresponding entropy of the resulting image of the absolute difference ($G_i$), which it is given by, 
\begin{eqnarray}\label{entropy3}
E_{i}=-\sum\limits_{j\in I_k}p_{j}\log_2{p_{j}}, &i=1,2,3,\ldots \nonumber
\end{eqnarray}

where $I_k=\left\lbrace j\in\left[0,2^B-1\right]:p_j\neq 0\right\rbrace $.

As the mean shift is not idempotent and one is in presence of an iterative process, of iteration to iteration the component elements (including the background) of the resulting image will be more homogeneous.
\begin{remark}
One should remember that the mean shift has a behavior of low pass filter
\end{remark}
Taking in consideration, that the data set (pixels) $\left\lbrace x_i \right\rbrace_{1,2,3,\ldots}$  have $n$ finite cardinality and that the succession of images $G_i$ are bound in the $(k+i)^{th}$ iteration, then $I_{k+i}$ image will be closer to the $I_{k+(i-1)}$ image. Therefore, due to the subtraction process $G_i \rightarrow 0$. This means that the histograms (density function) of these images will have a peak at the zero level.
 
So, 
$$\lim_{Ite\rightarrow \infty} G_{i}\rightarrow 0$$
(where Ite is the number of iterations). Therefore, if $p_i=\frac{k_i}{n\times m}$, where $k_i$ is the frequency of occurrence of $i$ gray level and $(n, m)$ are respectively the quantities of rows and columns of image; then 
\begin{align*}
k_0=n\times m\ \mbox{and}\ p_0=1 \Rightarrow \log(p_0=1)=0 \Rightarrow E_0=0.
\end{align*}

If $E_0$ is considered as the result of adding in expression (\ref{entropy2}), one arrives to a contradiction.

Proof of the first part. In \cite{Dominguez11,Comaniciu00} the convergence of the mean shift for norms $l_2$ and $l_{\infty}$ were respectively proven.

In the case of the MSHIA the convergence is interpreted when the resulting entropy of the absolute difference between the iteration image and the following falls below the stopping threshold. Taking into consideration the proof of the last part of the theorem becomes evident; whatever the chosen threshold the proof of the first part.

\end{document}